\documentclass[journal,onecolumn]{IEEEtran}

\usepackage{amsmath,amsfonts}
\usepackage{algorithmic}
\usepackage{algorithm}
\usepackage{array}
\usepackage[caption=false,font=normalsize,labelfont=sf,textfont=sf]{subfig}
\usepackage{textcomp}
\usepackage{stfloats}
\usepackage{url}
\usepackage{verbatim}
\usepackage{graphicx}
\usepackage{cite}
\usepackage[colorlinks,
            linkcolor=blue,
            anchorcolor=blue,
            citecolor=blue]{hyperref}
\usepackage[T1]{fontenc}

\usepackage{subfig}

\usepackage{booktabs}
\usepackage{multirow}
\usepackage{makecell}

\begin{document}

\title{Deformable Feature Alignment and Refinement for Moving Infrared Dim-small Target Detection}

\author{Dengyan~Luo, Yanping~Xiang,
Hu~Wang, Luping~Ji, ~\IEEEmembership{Member,~IEEE},  Shuai~Li, ~\IEEEmembership{Senior Member,~IEEE,} and Mao~Ye$^*$, ~\IEEEmembership{Senior Member,~IEEE}

\thanks{This work was supported in part by the National Natural Science Foundation of China (62276048) and Chengdu Science and Technology Projects (2023-YF06-00009-HZ).}
\thanks{Dengyan Luo, Yanping Xiang, Hu Wang, Luping Ji and Mao Ye are with the School of Computer Science and Engineering, University of Electronic Science and
Technology of China, Chengdu 611731, P.R. China (e-mail: dengyanluo@126.com;
xiangyp@uestc.edu.cn;  wanghu0833cv@gmail.com; jiluping@uestc.edu.cn;
cvlab.uestc@gmail.com).}
\thanks{Shuai Li is with the School of Control Science and Engineering, Shandong University, Jinan 250000, P.R. China (e-mail:shuaili@sdu.edu.cn).}
\thanks{*corresponding author}}


\markboth{}%
{Shell \MakeLowercase{\textit{et al.}}: A Sample Article Using IEEEtran.cls for IEEE Journals}



\maketitle

\begin{abstract}
The detection of moving infrared dim-small targets has been a challenging and prevalent research topic.
The current state-of-the-art methods are mainly based on ConvLSTM to aggregate information from adjacent frames to facilitate the detection of the current frame.
However, these methods  implicitly utilize motion information only in the training stage and fail to explicitly explore motion compensation, resulting
in poor performance in the case of a video sequence including
large motion.
In this paper, we propose a {\bf\it Deformable Feature Alignment and Refinement (DFAR)} method
based on deformable convolution to explicitly use motion context in both the training and inference stages.
Specifically, a Temporal Deformable Alignment (TDA) module based on the designed Dilated Convolution Attention Fusion (DCAF) block is developed to explicitly align the adjacent frames with the current frame at the feature level.
Then, the feature refinement module adaptively fuses the aligned features and further aggregates useful spatio-temporal information by means of the proposed Attention-guided Deformable Fusion (AGDF) block.
In addition, to improve the alignment of adjacent frames with the current frame, we extend the traditional loss function by introducing a new motion compensation loss.
Extensive experimental results demonstrate that the proposed DFAR method achieves the state-of-the-art performance on two benchmark datasets including DAUB and IRDST.

\end{abstract}


\section{Introduction}
\IEEEPARstart{M}{oving} infrared dim-small target detection plays an important role in numerous practical applications, such as civil and military surveillance, because harsh external environmental conditions do not degrade the quality of infrared image \cite{goodall2016tasking}. 
Although video target detection \cite{8543221,9508142,deng2021minet,jiao2021new,10179249,10384353} and small object detection \cite{8839730,9305976,9409729,10518061} have been developed rapidly in recent years, there are still challenges in video infrared dim-small target detection due to the following reasons:
1) Due to the long imaging distances, the infrared target is extremely small compared to the background.
2) The low intensity makes it difficult to accurately extract the shape and texture information of targets from complex backgrounds.
3) In actual scenarios, sea surfaces, buildings, clouds, and other clutter can easily disturb or submerge small targets.
Therefore, it is necessary to study moving infrared dim-small target detection.

In the past few decades, researchers have performed many studies on infrared dim-small target detection. 
The early work is mainly based on traditional paradigms, such as background modeling \cite{bai2010analysis} and  data structure \cite{kong2021infrared,zhu2020tnlrs}.
Although traditional methods have achieved satisfactory performance, 
there are too many hand-crafted components, 
which makes their application impractical in complex scenarios.
In contrast, learning-based methods can use deep neural networks to globally optimize the model on a large number of training samples.
Therefore, learning-based methods are becoming increasingly popular, greatly promoting the development of infrared dim-small target detection.

According to the number of input frames used, existing learning-based methods can be categorized into {\it single-frame} based approaches and {\it multi-frame} based approaches.
The early works \cite{wang2019miss,9279305,wang2022interior,li2023dense,sun2023receptive,10508299} focused on {\it single-frame} target detection, that is,
only one frame is input into the network at a time.
Although they can improve detection performance to a certain extent, their performance is still limited due to ignoring the temporal information  implied in consecutive frames.
To explore the temporal information, {\it multi-frame} target detection methods
are proposed \cite{liu2021dim,zhang2023infrared,chen2024sstnet}.
This kind of scheme allows the network to not only model intra-frame spatial features, but also extract inter-frame temporal information of the target to enhance feature representation.

\begin{figure}[t]
  \centering
  \includegraphics[width=0.5\linewidth]{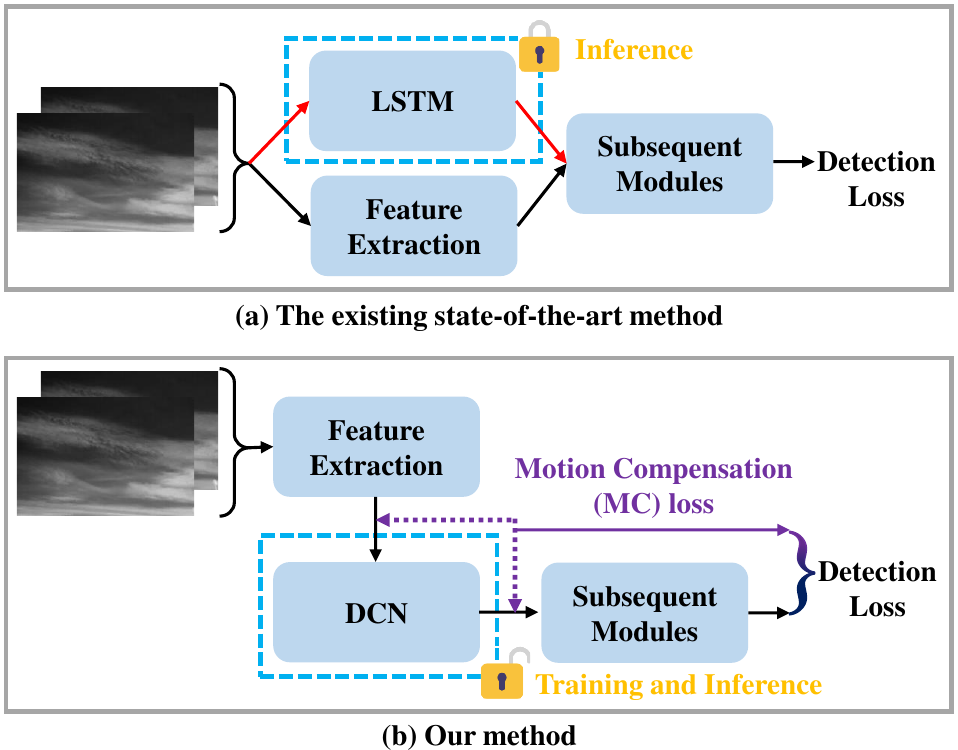}
  \caption{
  Illustrating the differences between our method and the state-of-the-art method SSTNet \cite{chen2024sstnet}. 
  The LSTM-based SSTNet implicitly aggregates the information from adjacent frames only in the training stage, while our method utilizes deformable convolution (DCN) to perform explicit inter-frame feature alignment in the training and inference stages, and applies  the introduced Motion Compensation (MC) loss to supervise the temporal alignment.
  }
  \label{compare}
\end{figure}

The existing multi-frame detection methods are mainly based on 3D
Convolutional structure \cite{liu2021dim,zhang2023infrared} or Convolutional Long Short-Term Memory (ConvLSTM) structure \cite{chen2024sstnet}.
However, these methods implicitly utilize motion information and fail to explicitly explore motion compensation, which limits the network's ability to model complex and large-scale motion.
Furthermore, the current state-of-the-art method SSTNet \cite{chen2024sstnet} only utilizes ConvLSTM to improve network optimization in the training stage, which results in the frame to be detected not actually using  the past-future frame  information in the inference stage, as illustrated in Fig. \ref{compare}. In addition, the models based on 3D convolution  or LSTM  are arduous to achieve a good trade-off
between computational cost and detection performance.

To alleviate the above issues, in this work, we propose a {\bf\it Deformable Feature Alignment and Refinement (DFAR)} method based on deformable convolution (DCN) \cite{Zhu_2019_CVPR} to explore motion context information simultaneously during training and inference stages, as presented in Fig. \ref{compare}. 
Our motivation stems from the observation that in moving infrared dim-small target detection, there exist many cases that the targets can not be detected in the current frame but are easier or harder to be perceived in some adjacent frames; and DCN has strong capabilities in modeling geometric transformations. 

Therefore, in order to adaptively utilize useful information from adjacent frames, there are two main parts are devised. On the one hand, a {\it Temporal Deformable Alignment (TDA)} module based on the designed  Dilated Convolution Attention Fusion (DCAF)
block is developed to explicitly align the adjacent frames with the current frame at the feature level. 
Specifically, the TDA module uses features from both the current frame and the adjacent frame to dynamically predict offsets of sampling convolution kernels. 
More specifically, the DCAF block uses channel attention to fuse multi-scale features extracted by multiple dilated convolutions, making the predicted offsets have an adaptive receptive field.
Then, dynamic kernels are applied on features from adjacent frames to employ the temporal alignment.
On the other hand, the {\it feature refinement} module adaptively fuses the 
feature of the current frame with the aligned adjacent features, and 
further aggregates effective spatio-temporal information through the proposed Attention-guided Deformable Fusion (AGDF) blocks.
In particular, the AGDF block adopts a pyramid offsets generation scheme and fuses multi-scale deformable offsets at the pixel level, 
which provides the model with the ability to implicitly model complex and large motions.
In addition, to improve the alignment effect, we introduce a new Motion Compensation (MC) loss $\mathcal{L}_{MC}$ by measuring the $L_1$ distance between the aligned adjacent features and the current frame feature.

The main contributions of this paper can be summarized as follows:
(1) A new {\it Deformable Feature Alignment and Refinement (DFAR)} method based on deformable convolution is proposed to mine temporal information implied in continuing frames, and effectively align the target frame with its adjacent frames through a designed Temporal Deformable Alignment (TDA) module respectively.
(2) We propose a feature refinement module to adaptively fuse the aligned adjacent features, 
and further explore valuable spatio-temporal details from the fused aligned features with the proposed Attention-guided Deformable Fusion (AGDF) blocks.
(3) A new Motion Compensation (MC) loss $\mathcal{L}_{MC}$ is proposed which is used to supervise the alignment of adjacent frames with the current frame.

Experimental results demonstrate that the proposed DFAR method achieves superior performance compared with the state-of-the-art methods in both quantitative and qualitative evaluations.   
The remainder of the paper is organized as follows:
Section \ref{Sec2} introduces the related works, including  single-frame based infrared small target detection schemes and multi-frame based infrared small target detection schemes.
Section \ref{Sec3} elaborates on the structure and details of the proposed method.
Section \ref{Sec4} shows comparative experiments and ablation studies on two benchmark datasets.
Finally, Section \ref{Sec6} draws conclusions.

\section{Related Works}
\label{Sec2}
\subsection{Single-frame Infrared Small Target Detection}
According to the number of frames used, infrared small target detection schemes can be classified into two categories: single-frame methods and multi-frame methods. Single-frame infrared small target detection aims to accurately detect targets in a single infrared image, which can be divided into two
categories: traditional methods \cite{bai2010analysis,kong2021infrared,zhu2020tnlrs,deshpande1999max,gao2013infrared,chen2013local,7795213,moradi2020fast,10168086} 
and deep learning-based methods 
\cite{liu2017image,dai2021asymmetric,9314219,10216342,10298041,lin2023ir,dai2023one,liu2023combining,10365220,chen2024tci,10553295,10497612}. 

Traditional methods can be  further categorized into background modeling, data structure and  target featuremethods. 
Background modeling methods estimate and suppress the background effect, such as  top-hat \cite{bai2010analysis} and  max-mean \cite{deshpande1999max}. 
Data structure methods are based on the sparsity of the target and the low rank of the background to separate the target and the background, such as TNLRS \cite{zhu2020tnlrs}  and IPI \cite{gao2013infrared}. 
Target feature methods are based on the feature differences between the target and its neighboring regions to detect the target, such as LCM \cite{chen2013local} and WSLCM \cite{moradi2020fast}.

Since the introduction of Multi-layer Perception (MLP) network into small target
detection by Liu et al. \cite{liu2017image}, 
deep-learning-based algorithms have received much research attention.
To take advantage of the Convolutional Neural Network (CNN), 
for example, 
Dai et al. \cite{dai2021asymmetric} proposed a ACM to aggregate low-level and deep-level features.
To unambiguously establish long-range contextual information, for example, 
Liu et al. \cite{10298041}  designed a method based on the transformer for infrared small target detection.
Later on,  Lin et al. \cite{lin2023ir} designed a IR-TransDet to integrate the benefits of the CNN and the transformer to properly extract global semantic information and features of small targets.
Although these methods have performed well in infrared image dim-small target detection, 
they are less effective in video infrared
dim-small target detection due to the ignorance of temporal information.

\subsection{Multi-frame Infrared Small Target Detection}

\begin{figure*}[t]
    \centering
    \includegraphics[width=0.99\textwidth]{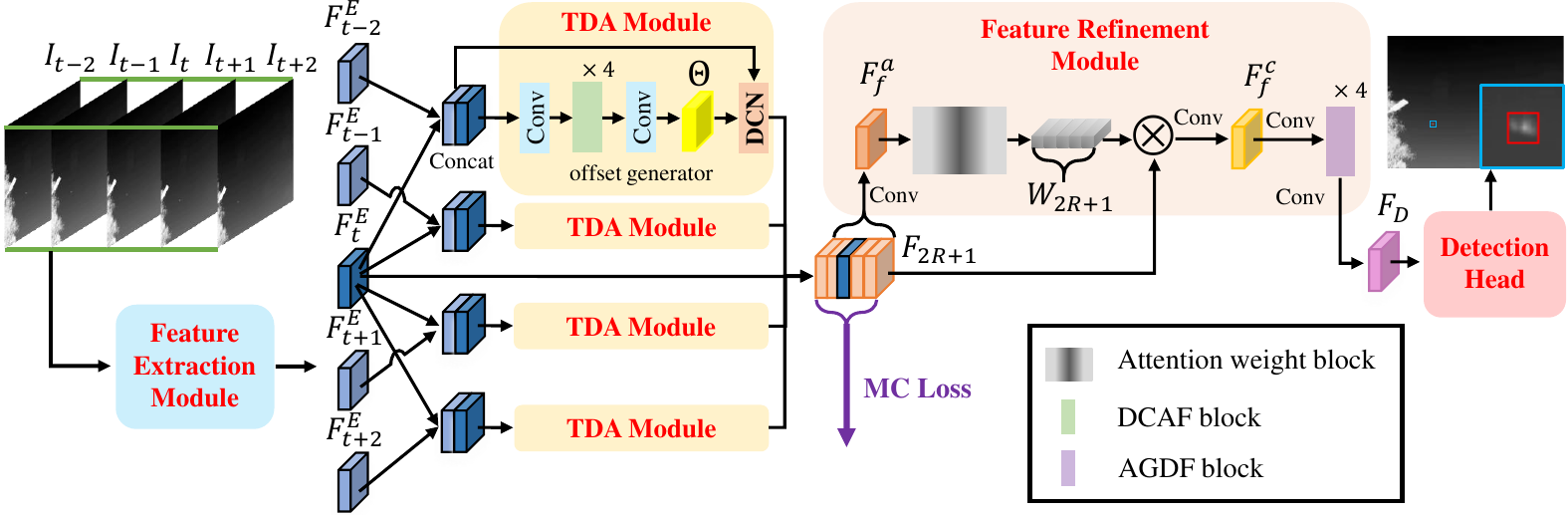}
    \caption{The framework of the proposed DFAR approach consists of four parts.
    (a) A feature extraction module is applied to extract the spatial information from the input clip $I_{[t-R,t+R]}$ and obtain the extracted features $F_{[t-R,t+R]}^{E}$.
    (b) The extracted visual features $F_{i}^{E}$ and 
$F_{t}^{E}$ are concatenated in the channel dimension and fed into the Temporal Deformable Alignment (TDA) module based on the  Dilated Convolution Attention Fusion (DCAF) blocks for alignment, $i \in[t-R,t+R] \text { and } i \neq t$.
 (c) The aligned features and the extracted target feature are input into the feature refinement module based on the attention weight block and the designed Attention-guided Deformable Fusion (AGDF) blocks to adaptively fuse and refine spatio-temporal information.
 (d) The refined feature $F_D$ is fed into the detection head module for calculating the detection loss.
 The network is optimized under the supervision of the traditional detection loss and the introduced Motion Compensation (MC) loss.
Herein, temporal radius $R = 2$.}
  \label{framework}
\end{figure*}

To simultaneously utilize spatio-temporal information, 
multi-frame detection methods have been proposed in recent years.
The early multi-frame methods \cite{sun2019infrared,kwan2020enhancing,9293143,wang2021infrared,liu2021nonconvex,wu2023infrared,10375328} are also non-intelligent learning, and they are mainly based on tensor optimization.
For instance, 
Sun et al. \cite{sun2019infrared}  developed a STTV-WNIPT in combination with spatial and temporal information to separate the target and background. 
Then, kwan et al. \cite{kwan2020enhancing} proposed to use optical flow to improve detection performance.
Later on, Wu et al. \cite{wu2023infrared} proposed to construct a 4-D spatio-temporal tensor and decompose it into a low-dimensional tensor.
However, these methods are heavily dependent on traditional priors and handcrafted features, 
resulting in poor detection performance in complex scenes such as clutter and noise.

To overcome the above weakness, learning-based multi-frame detection methods 
\cite{liu2021dim,zhang2023infrared,chen2024sstnet,du2021multiple,du2021spatial,yan2023stdmanet,li2023direction,10521471,tong2024st} are proposed.
For example, Du et al. \cite{du2021spatial} proposed a STFBD to use multiple frames for video infrared dim-small target detection.
Yan et al. \cite{yan2023stdmanet} then designed a STDMANet to explore temporal multi-scale features.
However, these methods directly 
concatenate multiple frames or features to construct spatio-temporal tensors resulting in rude motion information fusion.
Later on, Zhang et al. \cite{zhang2023infrared} utilized 3D CNN and traditional priors to mine motion information. 
Meanwhile,  Li et al. \cite{li2023direction} developed a DTUM based on 3D CNN to encode the motion direction into features and extract the motion information of targets.
After that, Tong et al. \cite{tong2024st} introduced a ST-Trans based on the transformer to learn the spatio-temporal dependencies between successive
frames of small infrared targets.
More recently, Chen et al. \cite{chen2024sstnet} proposed a  SSTNet based on ConvLSTM for multi-frame infrared small target detection.
Although the state-of-the-art performance has been achieved, 
in these works, motion information is modeled implicitly and fails to be compensated explicitly, resulting in poor performance.
Moreover, the models based on 3D CNN, transformer and LSTM tend to have heavy parameters and computation.
In addition, the SSTNet method does not utilize ConvLSTM in the inference stage but only in the training stage to aggregate past–current–future frames, which may also lead to overfitting in the training dataset.
In contrast, we directly incorporate deformable alignment into our module, allowing an explicit guidance during training and inference stages, thus achieving better performance and faster speed than the existing learning-based multi-frame methods.


\section{The Proposed Method}
\label{Sec3}
\subsection{Overview}
Given a video clip of $2R + 1$ consecutive frames 
$I_{[t-R,t+R]}$,
the middle frame $I_{t} \in \mathbb{R}^{C \times H \times W}$ is the target frame to be detected and the other frames are the reference frames.
Here, $R$ is the temporal radius (i.e., the number of input frames is $2 \times R+1$), $C$ refers to the channel number, and $H \times W$ denotes
the frame size.
Our goal is to improve detection performance by enabling the network to learn motion context features.
The overall structure of our DFAR method is shown in Fig. \ref{framework},
which consists of four modules:
a feature extraction module, 
a Temporal Deformable Alignment (TDA) module based on Dilated Convolution Attention Fusion (DCAF) blocks for feature alignment,
a feature refinement module based on the attention weight block and the proposed Attention-guided Deformable Fusion (AGDF) blocks, 
and a detection head module for target detection. 

As shown in Fig. \ref{framework}, 
firstly,
a feature extraction module  is applied to extract the spatial information for each frame from the input clip $I_{[t-R,t+R]}$ and obtain the extracted features $F_{[t-R,t+R]}^{E} \in \mathbb{R}^{c  \times h \times w}$, where $c$, $h$, and $w$ denote the channel, height, and width
of the feature $F_{i}^{E}$, respectively.
The feature extraction can be represented as:
\begin{equation}
  F_{[t-R,t+R]}^{E}=FE(I_{[t-R,t+R]}),
\end{equation}
where $FE(\cdot)$ denotes the feature extraction module, which is a three-level pyramid structure, and  each level contains a standard $3 \times 3$ convolutional layer with a stride of 1 and 
a $3 \times 3$ convolutional layer with a stride of 2 for downsampling. 
Then, each adjacent  frame feature $F_{i}^{E}$ enters the TDA module along with the current frame feature $F_{t}^{E}$ for temporal alignment:
\begin{equation}
  F_{i}^{A}=TDA(F_{i}^{E}, F_{t}^{E}), i \in[t-R,t+R] \text { and } i \neq t,
\end{equation}
where $F_{i}^{A}$ is the each aligned feature.
Furthermore, each aligned feature $F_{i}^{A}$ and the extracted target feature $F_{t}^{E}$ are input into the feature refinement module to further gather valuable spatio-temporal information:
\begin{equation}
  F_D=FR(F_{i}^{A}, F_{t}^{E}),  
\end{equation}
where $FR(\cdot)$ denotes the feature refinement module, and $F_D$ is the refined feature.
Finally, following SSTNet \cite{chen2024sstnet}, the refined feature $F_D$ is input to the detection head YOLOX \cite{ge2021yolox} for target detection.
It should be noted that for the first few frames and the last few frames in a video sequence, where the number of adjacent frames is less than $2 \times R$, we repeatedly pad it with the target frame until there are $2 \times R$ frames.

The details of the TDA module and the feature refinement module are explained in the following Sec. \ref{3.2} and Sec. \ref{3.3},
while the proposed Motion Compensation (MC) loss $\mathcal{L}_{MC}$ is presented in Sec. \ref{3.4}.

\begin{figure}[t]
  \centering
  \includegraphics[width=0.5\linewidth]{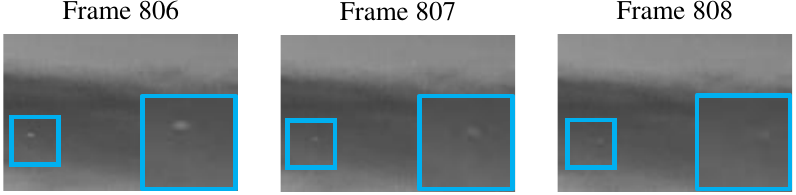}
  \caption{An example of moving infrared dim-small target. The target is more easily perceived in the adjacent frame 806 than in the current frame 807 and more difficult to be detected in the adjacent frame 808.}
  \label{align}
\end{figure}

\subsection{Temporal Deformable Alignment Module}
\label{3.2}

The motivation for inter-frame alignment comes from our observation that in moving
infrared dim-small target detection task, 
there are many situations where it is not easy to detect the target in the current frame, 
but the related target information exists in adjacent frames.
Fig. \ref{align} shows an example of this case, where the target in  frame 807 is difficult to be detected, whereas the target in  frame  806 is much more easily perceived.
And deformable convolution (DCN) has shown promising performance at capturing the motion cues of the targets in some 
low-level vision tasks like video super-resolution \cite{song2021multi} and video deraining \cite{yan2022feature}.
Therefore, we design a TDA module based on DCN to aggregate temporal information.

Concretely, firstly, the adjacent feature $F_{i}^{E}$ and the target feature $F_{t}^{E}$ are concatenated in the channel dimension and then passed through a $3\times3$ convolutional layer to make the number of channels consistent:
\begin{equation}
  F_{ti}=Conv([F_{t}, F_{i}]),
\end{equation}
where $[\cdot,\cdot]$ denotes the concatenation operation in the channel axis.
Then,  inspired by the residual dense network \cite{zhang2018residual}, 
we design a Dilated Convolution Attention Fusion (DCAF) block as the basic block to further integrate
the temporal information: 
\begin{equation}
  F_{ti}^{'}= (DCAF)^4( F_{ti} ) .
\end{equation}
The stacked blocks allow the network to have a large enough receptive field to aggregate information from distant spatial locations.
Finally, the aggregated feature $F_{ti}^{'}$ is input to a $3 \times 3$ convolutional layer to generate the corresponding deformable sampling parameters for alignment:
\begin{equation}
 \Theta=Conv(F_{ti}^{'}),
\end{equation}
where $\Theta \in \mathbb{R}^{d \times 2 K^{2} \times h \times w}$ is offset filed for the deformable convolutional kernel; $d$ and $K^{2}$ denote the deformable group and the kernel size of the deformable convolution, respectively.
Then, deformable convolution with the predicted offsets $\Theta$ is applied to the feature $F_{i}^{E}$ to get the aligned feature $F_{i}^{A}$:
\begin{equation}
  F_{i}^{A}(p)=\sum_{k=1}^{ K^{2} } \omega_{k} \cdot F_{i}^{E}(p+p_{k}+\Delta p_{k}),
  \label{eq3}
\end{equation}
where 
$p_{k}$ denotes the sampling grid with $K^{2}$ sampling locations, and $\omega_{k}$ represents the weights for each location $p$;
$\Delta p_{k}$ is the learnable offset for the $k$-th location, that is $\Theta = \{\Delta p_{k}\}$.
As the $p+p_{k}+\Delta p_{k}$ can be fractional, bilinear interpolation is adopted as in \cite{dai2017deformable}.
It should be noted that the above process only describes the prediction and application of offsets.
In fact, we also generate and apply masks for deformable convolution \cite{Zhu_2019_CVPR}.

\begin{figure}[t]
  \centering
  \includegraphics[width=0.5\linewidth]{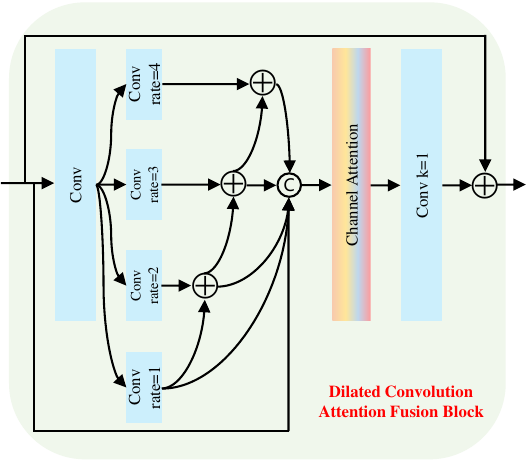}
  \caption{
  The structure of Dilated Convolution Attention Fusion (DCAF) block. It contains 4 dilated convolutions with a dilation rate from 1 to 4.
  With no special indication, 
  the kernel size and dilation rate of the convolutional layer are set to $3 \times 3$ and 1, respectively.
  }
  \label{DCAF_block}
\end{figure}

The structure of the DCAF block is depicted in Fig. \ref{DCAF_block}. 
In detail, to reduce 
computational cost while ensuring detection performance, we first use a $3 \times 3$ convolutional layer to halve the number of channels.
Then, 4 dilated convolutions with a dilation rate from 1 to 4 are used to obtain feature maps with different receptive fields.
These feature maps are hierarchically added before concatenating them with the original input feature to acquire an effective receptive field.
After that, the channel attention \cite{zhang2018image} and a  $1 \times 1$ convolutional layer are utilized to fuse the concatenated multi-scale feature and restore the input channels of the DCAF block. 
The channel attention mechanism can play a selection role for different receptive fields, which enables our TDA module to adaptively model different degrees of motion between frames.
Finally, the local skip connection with residual scaling is applied to complete our DCAF block.

\begin{figure}[t]
  \centering
  \includegraphics[width=0.5\linewidth]{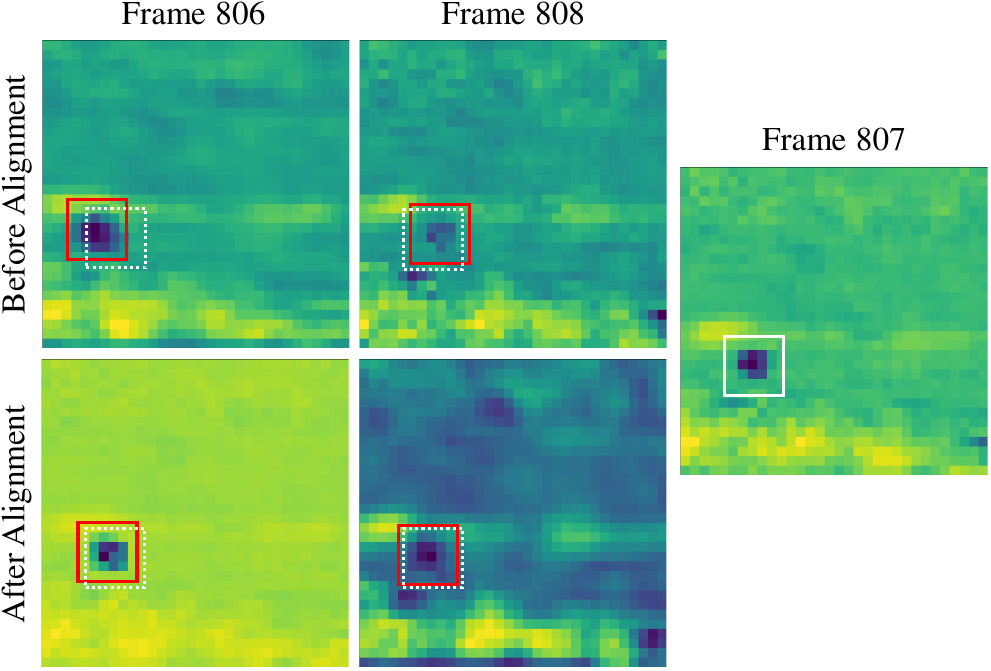}
  \caption{An example of visualizing feature maps. 
  The feature maps are obtained by averaging all corresponding channel features.
  The white area  is the anchor area to be aligned, 
  and the red areas indicate the target areas.
  After alignment, the target features in adjacent frames are closer to the target features in the detected frame and become more easily perceived and utilized. 
  \textit{Zoom in for the best view.}
  }
  \label{align_feature}
\end{figure}

{\bf Remark.}
Although optical flow can also be used for explicit temporal alignment, optical flow estimation only predicts an offset for each coordinate, and this single-coordinate single-offset mechanism severely restricts the modeling ability in more complex scenarios.
In addition, per-pixel motion estimation often suffers a heavy computational load.
However, our method aligns the target frame with adjacent frames at the feature level, 
which makes the network have strong capability and flexibility to handle various motion conditions in temporal scenes.
We visualize the feature maps in Fig. \ref{align_feature} to intuitively illustrate the effectiveness of the TDA module.
We can see from  Fig. \ref{align_feature} that after alignment, 
the target feature in frame 806 is closer to the target feature in frame 807.
This makes it easier for the network to perceive the target with the help of information from adjacent frames.
At the same time, the object feature in frame 808 that is prone to introducing new artifacts becomes more easily perceived.

\subsection{Feature Refinement Module}
\label{3.3}

Although we explicitly explore motion information, ineffective alignment can lead to worse detection results if adjacent frames are too blurry.
As shown in Fig. \ref{align}, the target is harder to be detected in frame 808 than in frame 807, and the alignment may introduce new artifacts, making the target more susceptible to disturbance. Therefore, we design a feature refinement module to adaptively fuse and improve useful temporal information from the aligned features.

As shown in Fig. \ref{framework}, we first aggregate the aligned features $ F_{i}^{A}$ with the extracted feature $F_{t}^{E}$ via a concatenation operation followed
by a 1$\times$1 convolutional layer to derive the fused alignment feature $F_{f}^{a}$:
\begin{equation}
  F_{f}^{a}=Conv([F_{t-R}^{A}, \ldots, F_{t-1}^{A}, F_{t}^{E}, F_{t+1}^{A}, \ldots, F_{t+R}^{A}]).
\end{equation}
Then, a global average pooling operation and two $1 \times 1$ convolutional layers are adopted to generate the attention weights  $W_{2R+1}$:
\begin{equation}
W_{2R+1}=Conv_{2R+1}(Conv(G A P(F_{f}^{a}))),
\end{equation}
where $GAP(\cdot)$ represents the global average pooling operation, and
$Conv_{2R+1}(\cdot)$ denotes the convolution operation required to generate weights $W_{2R+1}$ for the features to be aggregated.
After that, the adaptive fusion weights $W_{2R+1}$ are element-wise multiplied with the features $F_{2R+1}= \{F_{t-R}^{A}, \ldots, F_{t-1}^{A}, F_{t}^{E}, F_{t+1}^{A}, \ldots, F_{t+R}^{A}\}$:
\begin{equation}
\widetilde{F}_{2R+1}=F_{2R+1} \otimes W_{2R+1},
\end{equation}
where $\otimes$ refers to element-wise multiplication operation. 
Finally, the generated modulated features $\widetilde{F}_{2R+1}$ are concatenated
in the channel dimension and then via a 1$\times$1 bottleneck convolution to obtain the fused coarse feature $F_{f}^{c}$:
\begin{equation}
F_{f}^{c}=Conv([\widetilde{F}_{t-R}, \cdots, \widetilde{F}_{t-1}, \widetilde{F}_{t}, \widetilde{F}_{t+1}, \cdots, \widetilde{F}_{t+R}]).
\end{equation}

The obtained feature $F_{f}^{c}$ contains both intra-frame spatial relation and inter-frame temporal information, and the temporal dependence among frames is concatenated along the channel dimension.  
Therefore, we need to design an effective fusion structure to enable the network to dynamically aggregate spatio-temporal information.
Fortunately, the attention mechanism  can highlight the important information and suppress
the less useful information. 
Thus, we propose a spatial attention and channel attention guided AGDF block to enable the network 
to concentrate on more valuable information and enhance discriminative learning ability.

\begin{figure}[t]
  \centering
  \includegraphics[width=0.5\linewidth]{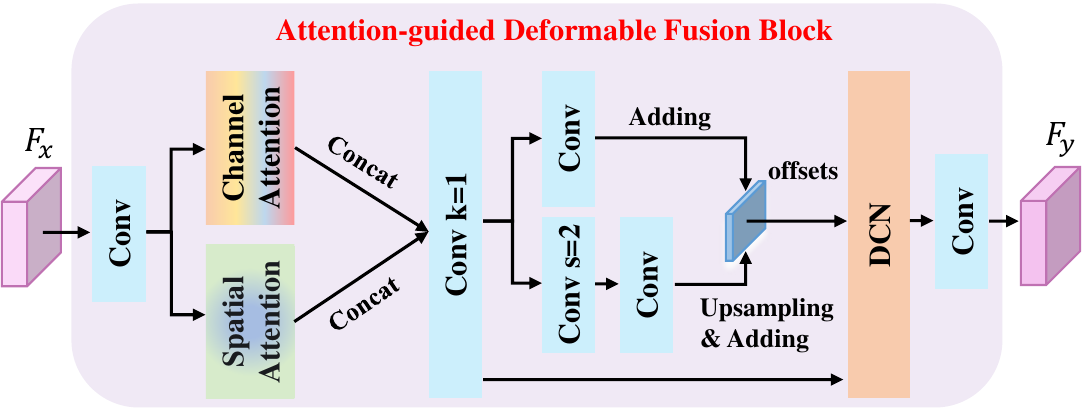}
  \caption{
  The structure of Attention-guided Deformable Fusion (AGDF) block. 
  With no special indication, 
  the kernel size and stride of the convolutional layer are set to $3 \times 3$ and 1, respectively.
  }
  \label{AGDF_block}
\end{figure}

The structure of the AGDF block is illustrated in Fig. \ref{AGDF_block}.
Assuming that the input feature map of the AGDF block is $F_x$, 
we first use a $3 \times 3$ convolutional layer to halve the number of channels for low computational cost:
\begin{equation}
F_{x}^{'}=Conv(F_x).
\end{equation}
Then, channel attention and  spatial attention \cite{woo2018cbam} are applied to aggregate spatial information and temporal dependence respectively:
\begin{equation}
F_{x}^{s}=SA(F_{x}^{'}),
\end{equation}
\begin{equation}
F_{x}^{c}=CA(F_{x}^{'}),
\end{equation}
where $SA(\cdot)$ and $CA(\cdot)$ represent spatial attention and channel attention operations, respectively.
Furthermore, a $1 \times 1$ convolutional layer is  adopted to combine two branch features and make the number of channels consistent:
\begin{equation}
F_{x}^{sc}=Conv([F_{x}^{s}, F_{x}^{c}]).
\end{equation}
After that, we utilize a multi-scale structure to predict offsets and perform deformable fusion.
Specifically, we use a strided convolution to downsample $F_{x}^{sc}$ by
a factor of 2 and predict offsets at two different scales:
\begin{equation}
\Theta_{1}^{sc} =Conv(F_{x}^{sc}),
\end{equation}
\begin{equation}
\Theta_{2}^{sc} =Conv( SConv(F_{x}^{sc}) ).
\end{equation}
Then, deformable convolution with the fused multi-scale offsets are applied to the mixed spatio-temporal feature $F_{x}^{sc}$ to further fuse spatio-temporal information:
\begin{equation}
F_{x}^{f} =DCN( F_{x}^{sc}, (\Theta_{2}^{sc})^{\uparrow 2} + \Theta_{1}^{sc} ),
\end{equation}
where $(\cdot)^{\uparrow 2}$ refers to upscaling by a factor of 2.
The deformable convolution with pyramid generated offsets allows the AGDF block to have a larger and adaptive receptive field to aggregate information.
Finally, a $3 \times 3$ convolutional layer is used to increase the number of channels to ensure consistent input and output channels for the  AGDF block:
\begin{equation}
F_{y} =Conv( F_{x}^{f} ).
\end{equation}

{\bf Remark.} 
Although the stacked convolutional layers or multi-scale convolutional structures can also increase the receptive field of the network, gradient vanishing/exploding and degradation problems may be caused. 
In contrast, our AGDF block employs deformable convolution with joint predicted offsets 
to model spatio-temporal information from the fused two-branch features, 
leading to more efficient use of both intra-frame and inter-frame information.
Although the three-level pyramid structure may bring more performance improvement, the accompanying increase in computational cost is unacceptable. 
Our AGDF block reaches a balance between the performance and computational efficiency by the designed two-level pyramid structure.

    

\subsection{The Loss Function}
\label{3.4}
Although the deformable alignment has the potential to capture motion context and align $F_{i}^{E}$ with $F_{t}^{E}$, it is difficult to train deformable convolution without a supervision loss \cite{Chan2021UnderstandingDA}. 
Training instability regularly leads to offset overflow, degrading the final performance.
Therefore, to improve the temporal alignment, we introduce a new Motion Compensation (MC) loss $\mathcal{L}_{MC}$  as follows: 
\begin{equation}
\mathcal{L}_{M C}=\sum_{i=t-R, \neq t}^{t+R} \mathcal{L}_1\left(F_i^A, F_t^E\right),
\end{equation}
where $\mathcal{L}_1(\cdot,\cdot)$ represents $L_1$ loss.
Then, the total loss function is formulated as:
\begin{equation}
\mathcal{L}=\lambda \mathcal{L}_{r e g}+\mathcal{L}_{c l s}+\mathcal{L}_{o b j}+\eta \mathcal{L}_{M C},
\label{loss}
\end{equation}
where $\mathcal{L}_{r e g}$ is a regression loss, $\mathcal{L}_{c l s}$ denotes a classification loss, and $\mathcal{L}_{o b j}$ refers to an IoU loss; 
$\lambda$ and $\eta$ are two hyper-parameters to balance loss terms. Here, following the setting of YOLOX \cite{ge2021yolox}, $\lambda$ is fixed to 5.

\section{Experiments}
\label{Sec4}

\subsection{Datasets and Quantitative Evaluation Metrics}
\textit{Datasets.}
Following \cite{chen2024sstnet}, 
we conduct extensive experiments on 
two moving infrared dim-small target detection datasets, 
i.e., DAUB \cite{hui2019dataset} and IRDST \cite{sun2023receptive}.
The DAUB dataset consists of 10 training video sequences with a total of 8982 frames and 7 test video sequences with a total of 4795 frames.
The IRDST dataset contains 42 training video sequences with a total of 20398 frames and 43 test video sequences with a total of 20258 frames.

\textit{Quantitative Evaluation Metrics.}
To evaluate the detection performance,
four standard evaluation metrics for infrared dim-small target detection are adopted, that is Precision (Pr), Recall (Re), $F1$ score and the average precision (e.g., mAP$_{50}$, the mean average precision with the IoU threshold of 0.5).
These quantitative evaluation metrics are defined as follows:
\begin{equation}
\text { Precision }=\frac{\mathrm{TP}}{\mathrm{TP}+\mathrm{FP}},
\end{equation}
\begin{equation}
\text { Recall }=\frac{\mathrm{TP}}{\mathrm{TP}+\mathrm{FN}},
\end{equation}
\begin{equation}
F1 = \frac{ 2 \times \text { Precision } \times \text { Recall } } { \text { Precision } + \text { Recall } },
\end{equation}
where TP, FP, and FN denote the number of correct predictions
(true positives), false detections
 (false positives), and missing targets (false negatives), respectively.
The $F1$ score is a reliable and comprehensive evaluation metric on Pr and Re.

\subsection{Implementation Details}
\textit{Network settings.}
The convolutional layer in the feature extraction module has 48 filters (except for the last layer which has 64 filters). 
The kernel size of all deformable convolutions in the network is set to $3 \times 3$; 
the number of deformable groups in the TDA module and the AGDF block is set to 8 and 32, respectively.
There are 4 DCAF blocks and 4 AGDF blocks in the TDA module and feature refinement module, respectively.
For the other settings in the proposed DFAR method have been described in Section \ref{Sec3}.

\textit{Model training.}
To have a fair comparison, following \cite{chen2024sstnet}, 
the number of input frames is set to 5 (i.e., temporal radius $R = 2$).
In the training process, the input frames are reshaped into $544 \times 544$, and the batch size is set to 4.
The model is trained by Adam optimizer with $\beta_{1}=0.9$, $\beta_{2}=0.999$ and $\varepsilon=1 \times 10^{-8}$ for 20  epochs.
The learning rate is initially set to $1 \times 10^{-4}$ and retained throughout training.
For the hyper-parameters in equation (\ref{loss}), following YOLOX \cite{ge2021yolox}, $\lambda$ is set to 5; 
$\eta$ is selected by the grid search method and set to 1.
The proposed model is implemented using PyTorch, and trained on a NVIDIA GeForce RTX 3090 GPU.

\begin{table*}[htbp]
  \centering
  \caption{Overall Performance Comparison in Terms of mAP$_{50}$, Precision (Pr), Recall (Re) and $F1$ Score on the DAUB and IRDST Datasets.
  \textcolor{red}{Red} and \textcolor{blue}{Blue} Colors Indicate the Best and the Second-Best Performance, Respectively.}
    \setlength{\tabcolsep}{1.35mm}

    \begin{tabular}{c|c|c|cccc|cccc}
    \hline
    \multirow{2}{*}{\textbf{Scheme }} & \multirow{2}{*}{\textbf{Methods }} & \multirow{2}{*}{\textbf{Publication}} & \multicolumn{4}{c|}{\textbf{DAUB}} & \multicolumn{4}{c}{\textbf{IRDST}} \\
\cline{4-11}
&       &       & \textbf{mAP$_{50}$ (\%)} & \textbf{Pr (\%)} & \textbf{Re (\%)} & \textbf{F1 (\%)} & \textbf{mAP$_{50}$ (\%)} & \textbf{Pr (\%)} & \textbf{Re (\%)} & \textbf{F1 (\%)} \\
    \hline
    \multicolumn{1}{c|}{\multirow{12}[2]{*}{\textbf{  \thead{single-frame \\ based detection} }}} 
          & ACM \cite{dai2021asymmetric}   & WACV 2021 & 72.30 & 76.84 & 95.31 & 85.09 & 67.74 & 81.44 & 84.01 & 82.71 \\
          & RISTD \cite{hou2021ristdnet} & IEEE  GRSL 2022 & 82.73 & 88.54 & 94.41 & 91.38 & 78.92 & 86.56 & \textcolor{blue}{92.63} & 89.49 \\
          & ISNet \cite{zhang2022isnet} & CVPR 2022 & 83.43 & 88.64 & 95.04 & 91.73 & 75.00 & 87.78 & 86.81 & 87.29 \\
          & UIUNet \cite{wu2022uiu} & IEEE TIP 2022 & 88.23 & 94.63 & 94.79 & 94.71 & 70.96 & 87.28 & 82.08 & 84.60 \\
          & SANet \cite{zhu2023sanet} & ICASSP 2023 & 87.90 & 94.14 & 94.22 & 94.18 & 77.98 & 85.42 & 92.13 & 88.64 \\
          & AGPCNet \cite{zhang2023attention} & IEEE TAES 2023 & 73.08 & 78.49 & 94.41 & 85.72 & 73.86 & 83.77 & 89.18 & 86.39 \\
          & RDIAN \cite{sun2023receptive} & IEEE TRGS 2023 & 83.69 & 90.55 & 93.37 & 91.94 & 71.99 & 84.41 & 86.48 & 85.43 \\
          & DNANet \cite{li2023dense} & IEEE TIP 2023 & 89.24 & 95.66 & 94.83 & 95.24 & 76.84 & 90.08 & 86.81 & 88.42 \\
          & SIRST5K \cite{10496142} & IEEE TGRS 2024 & 88.45 & 94.48 & 94.97 & 94.72 & 72.64 & 86.17 & 85.65 & 85.91 \\
          & MSHNet \cite{liu2024infrared} & CVPR 2024 & 89.23 & 97.27 & 92.26 & 94.70 & 78.50 & 88.89 & 89.63 & 89.26 \\
          & RPCANet \cite{wu2024rpcanet} & WACV 2024 & 85.75 & 89.12 & 97.58 & 93.16 & 73.29 & 85.02 & 87.13 & 86.06 \\
          & SCTransNet \cite{10486932} & IEEE TGRS 2024 & 88.26 & 93.50 & 95.50 & 94.53 & 78.27 & 89.67 & 88.43 & 89.05 \\
    \midrule
    \multicolumn{1}{c|}{\multirow{3}[2]{*}{\textbf{ \thead{multi-frame \\ based detection} }}} 
          & DTUM \cite{li2023direction} & IEEE TNNLS 2023 & 88.24 & 95.15 & 93.60 & 94.37 & 80.98 & 90.62 & 90.46 & 90.54 \\
          & SSTNet \cite{chen2024sstnet} & IEEE TGRS 2024 & \textcolor{blue}{94.33} & \textcolor{blue}{97.77} & \textcolor{blue}{97.91} & \textcolor{blue}{97.84} & \textcolor{blue}{83.25} & \textcolor{blue}{91.13} & 92.24 & \textcolor{blue}{91.68} \\
          & \textbf{DFAR (Ours)} & -     & \textcolor{red}{96.56} & \textcolor{red}{98.82} & \textcolor{red}{98.62} & \textcolor{red}{98.72} & \textcolor{red}{89.88} & \textcolor{red}{95.91} & \textcolor{red}{94.66} & \textcolor{red}{95.28} \\
    \bottomrule
    \end{tabular}%

  \label{Comparisons}%
\end{table*}%

\begin{figure}[t]
  \centering
  \includegraphics[width=0.5\linewidth]{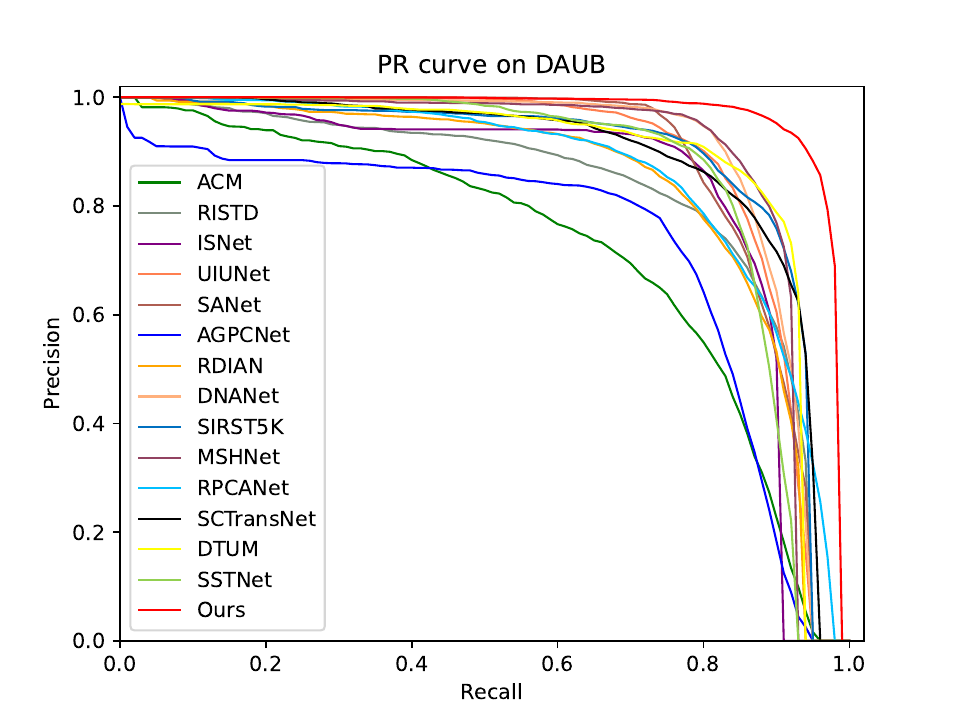}
  \caption{PR curve comparison on the DAUB dataset. The larger the area under the curve, the better the method.}
  \label{p-r-DAUB}
\end{figure}

\subsection{Comparisons with State-of-The-Art Methods}
To demonstrate the advantage of our method, we compare our
method with various learning-based infrared dim-small target detection approaches  whose codes are available,
including 12 single-frame based detection approaches: 
ACM \cite{dai2021asymmetric}, RISTD \cite{hou2021ristdnet}, ISNet \cite{zhang2022isnet}, UIUNet \cite{wu2022uiu}, SANet \cite{zhu2023sanet}, AGPCNet \cite{zhang2023attention}, RDIAN \cite{sun2023receptive}, DNANet \cite{li2023dense}, SIRST5K \cite{10496142}, MSHNet \cite{liu2024infrared}, RPCANet \cite{wu2024rpcanet} and SCTransNet \cite{10486932},
and 2 multi-frame based detection approaches: DTUM \cite{li2023direction} and SSTNet \cite{chen2024sstnet}.
The resolution of input frames for all comparison methods
in training and test is reshaped $544 \times 544$.
Since the datasets we used are based on bounding box annotations, 
to make the comparison as fair as possible,
for all single-frame detection methods (except the SANet method) and the DTUM method based on pixel-level segmentation, 
the detection head YOLOX \cite{ge2021yolox} is added to the output of their networks to generate bounding boxes.
Then, these methods are retrained on our training set.
For the SSTNet method, we directly run public training and test codes.

\begin{figure}[t]
  \centering
  \includegraphics[width=0.5\linewidth]{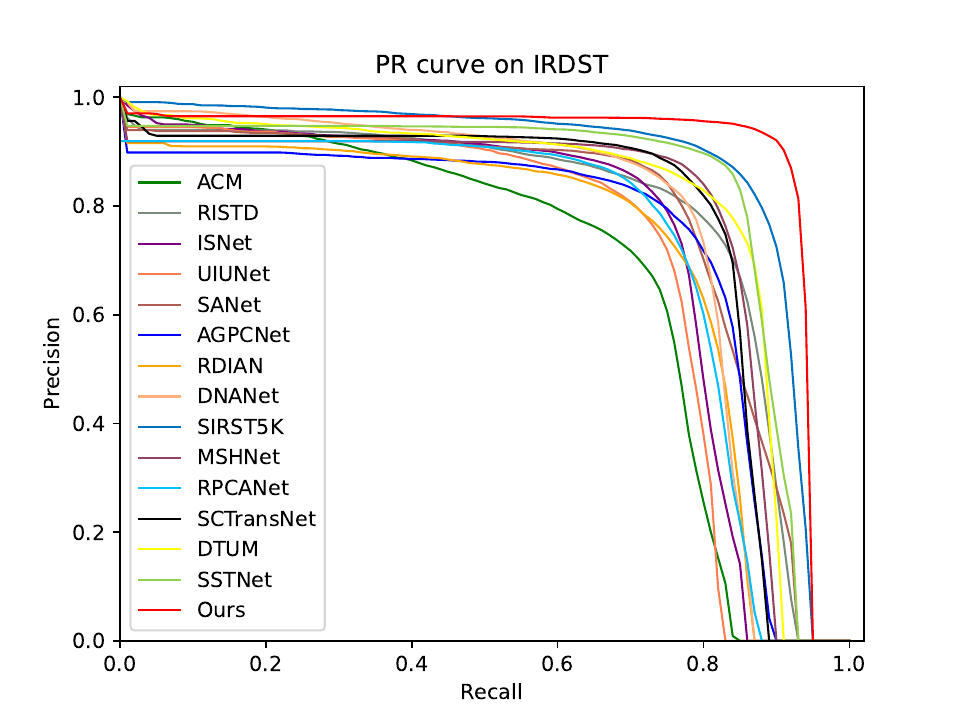}
  \caption{PR curve comparison on the IRDST dataset. The proposed DFAR approach (red curve)  obviously outperforms other methods.}
  \label{p-r-IRDST}
\end{figure}


\subsubsection{Overall Quantitative Comparison}
Table \ref{Comparisons} presents the quantitative comparison results on two datasets, averaged over all frames of each test sequence.
We can see that
our method achieves better results than all the compared methods on two datasets in terms of four standard evaluation metrics, i.e., mAP$_{50}$, Pr, Re and $F1$.

To be specific, on the DAUB dataset, 
the highest mAP$_{50}$ 96.56\% is reached by our method, which is 2.36\% higher than mAP$_{50}$ of SSTNet (94.33\%); 
the highest Pr 98.82\% is achieved by our method, which is 1.07\% higher than Pr of SSTNet (97.77\%); 
the highest Re 98.62\% is reached by our method,  0.73\% higher than Re of SSTNet (97.91\%); 
the highest $F1$ 98.72\% is achieved by our method, which is 0.90\% higher than $F1$ of SSTNet (97.84\%).

\begin{figure*}[t]
  \centering
  \includegraphics[width=0.825\linewidth]{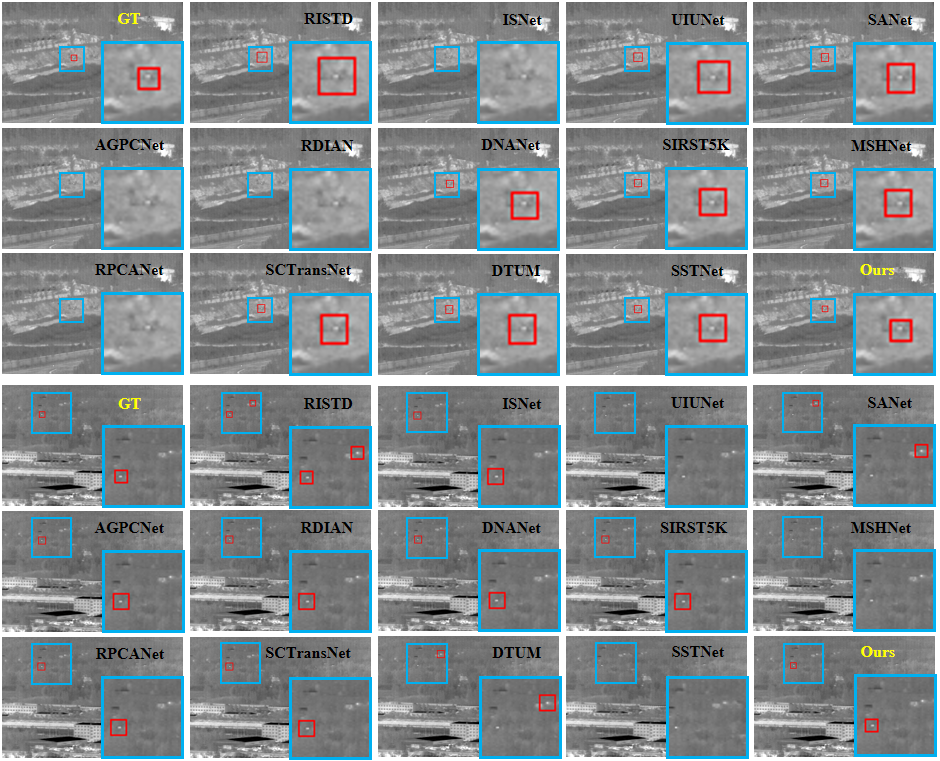}
  \caption{
Two groups of visualization comparisons on the DAUB dataset; GT: ground truth. Red box denotes a detected target, and
detection region is amplified (blue box).
}
  \label{Visualization_DAUB}
\end{figure*}

More specifically, on the IRDST dataset, 
the highest mAP$_{50}$ 89.88\% is reached by our method, which is 7.96\% higher than mAP$_{50}$ of SSTNet (83.25\%); 
the highest Pr 95.91\% is achieved by our method, which is 5.25\% higher than Pr of SSTNet (91.13\%); 
the highest Re 94.66\% is reached by our method, which is  2.62\% higher than Re of SSTNet (92.24\%); 
the highest $F1$ 95.28\% is achieved by our method, which is 3.93\% higher than $F1$ of SSTNet (91.68\%).
It can be seen that our method significantly outperforms the current state-of-the-art method SSTNet on the IRDST dataset. 
One possible reason is that the video sequences in the IRDST dataset have larger motions than those in the DAUB dataset, and it is difficult for LSTM-based SSTNet to implicitly aggregate inter-frame information. 
It demonstrates the robustness of our DFAR approach based on DCN explicit alignment in modeling complex and large motions.


\begin{table}[t]
  \centering
  \caption{
  Inference Complexity Comparison on the DAUB Dataset.
  For a Fair Comparison, All Methods are Retested on a NVIDIA GeForce RTX 3090.
  The Results are Reported by Model Parameters (Params), Floating-Point Operations (FLOPs) and Frame Per
Second (FPS). 
  The best Results are Marked in \textbf{Bold}.}
    \setlength{\tabcolsep}{0.5mm}

    \begin{tabular}{l|c|cc|rrr|c}
    \toprule
    \textbf{Methods} & \textbf{Frames} & \textbf{mAP$_{50}$}$^{\uparrow}$ & \textbf{F1}$^{\uparrow}$ & \textbf{Params}$^{\downarrow}$ & \textbf{FLOPs}$^{\downarrow}$ & \textbf{FPS}$^{\uparrow}$ & \textbf{PCR}$^{\uparrow}$ \\
    \midrule
    ACM \cite{dai2021asymmetric}  & 1     & 72.30 & 85.09 & 3.01M  & \textbf{28.17G} & \textbf{16.39} & \textbf{2.57} \\
    RISTD \cite{hou2021ristdnet} & 1     & 82.73 & 91.38 & 3.26M  & 92.95G  & 9.91  & 0.89  \\
    ISNet \cite{zhang2022isnet} & 1     & 83.43 & 91.73 & 3.48M  & 300.64G  & 6.44  & 0.28  \\
    UIUNet \cite{wu2022uiu} & 1     & 88.23 & 94.71 & 53.03M  & 515.83G  & 2.10  & 0.17  \\
    SANet \cite{zhu2023sanet} & 1     & 87.90 & 94.18 & 12.40M  & 47.46G  & 6.26  & 1.85  \\
    AGPCNet \cite{zhang2023attention} & 1     & 73.08 & 85.72 & 14.85M  & 413.61G  & 3.17  & 0.18  \\
    RDIAN \cite{sun2023receptive} & 1     & 83.69 & 91.94 & \textbf{2.71M} & 57.37G  & 13.91  & 1.46  \\
    DNANet \cite{li2023dense} & 1     & 89.24 & 95.24 & 7.19M  & 152.58G  & 3.47  & 0.58  \\
    SIRST5K \cite{10496142} & 1     & 88.45 & 94.72 & 11.28M  & 204.88G  & 6.64  & 0.43  \\
    MSHNet \cite{liu2024infrared} & 1     & 89.23 & 94.70 & 6.56M  & 78.77G  & 12.42  & 1.13  \\
    RPCANet \cite{wu2024rpcanet} & 1     & 85.75 & 93.16 & 3.18M  & 432.27G  & 13.13  & 0.20  \\
    SCTransNet \cite{10486932} & 1     & 88.26 & 94.53 & 13.68M  & 115.34G  & 8.12  & 0.77  \\
    \midrule
    DTUM  \cite{li2023direction} & 5     & 88.24 & 94.37 & 2.79M  & 117.10G  & 7.58  & 0.75  \\
    SSTNet \cite{chen2024sstnet} & 5     & 94.33 & 97.84 & 11.95M  & 139.53G  & 6.71  & 0.68  \\
    \textbf{Ours} & 5     & \textbf{96.56} & \textbf{98.72} & 7.53M  & 100.87G  & 7.60  & 0.96  \\
    \bottomrule
    \end{tabular}%

  \label{Speed_parameter}%
\end{table}%

\subsubsection{Precision–recall (PR) Curve Comparison}
To comprehensively evaluate performance, we further evaluate the PR curves of all approaches on the DAUB and IRDST datasets. 
The larger the area under the curve, the better the method.
We can see from Figs. \ref{p-r-DAUB}  and \ref{p-r-IRDST} that our approach is obviously superior to other comparative methods on two datasets. 
In general, the proposed DFAR approach achieves a better balance of precision and
recall than other methods.

\subsubsection{Inference Complexity Comparison}
We use  Frame Per Second (FPS), Floating-point Operations (FLOPs) and model parameters to compare inference complexity.
As shown in Table \ref{Speed_parameter}, 
the proposed DFAR approach is better than the state-of-the-art LSTM-based method SSTNet in terms of both inference speed and model size.
In addition, our method has a moderate Performance–cost Ratio (PCR; i.e., mAP$_{50}$/FLOPs).
Particularly, the PCR of our method on the DAUB dataset is 0.96, which is 41.18\% higher than that of SSTNet (0.68).
Although ACM, RISTD, RDIAN and MSHNet methods have smaller model sizes, FLOPs and faster inference speed, by checking Table \ref{Comparisons} and Table \ref{Speed_parameter}, we can see  that our method achieves a good complexity-performance trade-off.


\subsubsection{Visualization Comparison}
\begin{figure*}[t]
  \centering
  \includegraphics[width=0.824\linewidth]{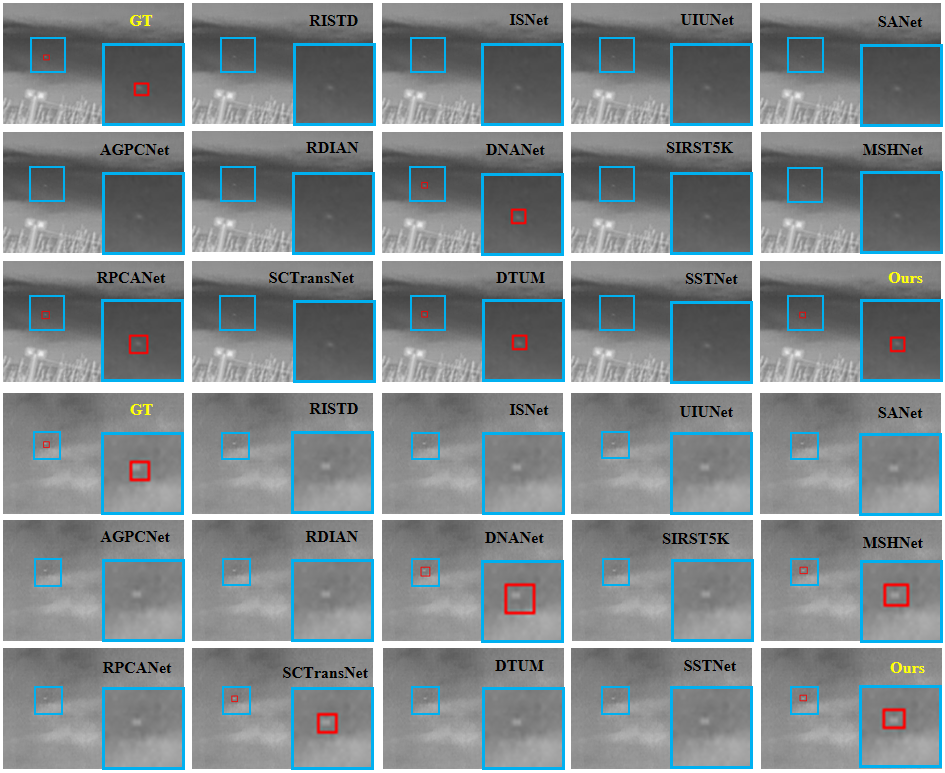}
  \caption{
Two groups of visualization comparisons on the IRDST dataset; GT: ground truth. Red box denotes a detected target, and
detection region is amplified (blue box).
}
  \label{Visualization_IRDST}
\end{figure*}

In Fig. \ref{Visualization_DAUB} and Fig. \ref{Visualization_IRDST}, we present two groups of detection results for visual comparison on the two datasets, respectively.
We can see from Figs. \ref{Visualization_DAUB} and \ref{Visualization_IRDST} that 
our method can often accurately detect moving dim-small targets, 
while other methods usually result in missed or false detections.

Specifically, on the DAUB dataset, as shown in Fig. \ref{Visualization_DAUB}, our method precisely detects the target in the first group of visualization comparison (the first three rows). However, ISNet, AGPCNet, RDIAN  and RPCANet methods cause missed detection; the target bounding boxes generated by RISTD and UIUNet methods are not accurate and appear too large.
Furthermore, in the second group of comparison (the fourth to sixth rows), 
UIUNet, MSHNet and SSTNet methods occur missed detection; 
RISTD, SANet and  DTUM methods produce false detection, and RISTD method even detect two targets.

More specifically, on the IRDST dataset, as shown in Fig. \ref{Visualization_IRDST},
most methods fail to detect two groups of targets since the targets are difficult to perceive, while our method can still detect the targets correctly. 
In summary, these qualitative comparison results
verify the superiority of our method.

\begin{table*}[htbp]
  \centering
  \caption{Ablation Study of Proposed Temporal Deformable Alignment (TDA) Module, Motion Compensation Loss $\mathcal{L}_{MC}$ and Feature Refinement (FR) Module.}

    \begin{tabular}{ccc|cccc|cccc}
    \hline
    \multirow{2}{*}{\textbf{TDA}} & \multirow{2}{*}{\textbf{  $\boldsymbol {\mathcal{L}_{MC} }$ }} & \multirow{2}{*}{\textbf{FR}} & \multicolumn{4}{c|}{\textbf{DAUB}} & \multicolumn{4}{c}{\textbf{IRDST}} \\
\cline{4-11}          
&       &       & \textbf{mAP$_{50}$ (\%)} & \textbf{Pr (\%)} & \textbf{Re (\%)} & \textbf{F1 (\%)} & \textbf{mAP$_{50}$ (\%)} & \textbf{Pr (\%)} & \textbf{Re (\%)} & \textbf{F1 (\%)} \\
    \hline
    
    -     & -     & -     & 84.02  & 87.40  & 97.14  & 92.02  & 74.13  & 86.24  & 88.28  & 87.25  \\
    $\checkmark$     & -     & -    & 92.62  & 95.40  & 97.87  & 96.62  & 81.44  & 90.50  & 90.86  & 90.68  \\
    $\checkmark$     & $\checkmark$     & -      & 93.05  & 96.52  & 97.89  & 97.20  & 82.44  & 91.92  & 92.12  & 92.02  \\
    -     & -     & $\checkmark$     & 91.92  & 95.44  & 97.71  & 96.56  & 80.88  & 92.43  & 88.56  & 90.45  \\
    $\checkmark$     & -     & $\checkmark$     & 94.69  & 97.79  & 97.93  & 97.86  & 87.14  & 95.14  & 92.28  & 93.69  \\
    $\checkmark$     & $\checkmark$     & $\checkmark$     & \textbf{96.56} & \textbf{98.82} & \textbf{98.62} & \textbf{98.72} & \textbf{89.88} & \textbf{95.91} & \textbf{94.66} & \textbf{95.28} \\
    \bottomrule
    \end{tabular}%

  \label{Ablation}%
\end{table*}%

\subsection{Ablation Study}
\subsubsection{Effects of Different Assemblies}
\label{Sec5}
To evaluate the effectiveness of our TDA module, feature refinement (FR) module and motion compensation loss, 
we directly use a $3 \times 3$ convolutional layer with 320 filters to fuse the concatenated multi-frame features, and then input fused feature into the detection head as the baseline model.
Then, we insert different components into the baseline, and retrain these models with the same experimental settings. The experimental results are presented in Table \ref{Ablation}.

\textbf{Effectiveness of TDA module.}
We insert the TDA module before multi-frame feature fusion to evaluate the effectiveness of explicit alignment on features.
The results in the 1st and 2nd rows (or 4th and 5th rows) in Table \ref{Ablation} prove the
effectiveness of our proposed TDA module.
Specifically, the results in the 4th and 5th rows show that alignment is more
important for the IRDST dataset which contains more large
motion videos.
Additionally, to evaluate the effectiveness of the designed DCAF block, 
we use 8 3 × 3 convolutional layers with 64 filters to replace the DCAF blocks in the TDA module to keep the module parameters and computational cost essentially unchanged. 
The mAP$_{50}$ and $F$1 of the baseline model with the modified TDA module on the DAUB dataset are only 89.86 and 94.18, and on the IRDST dataset are only 78.24 and 88.83.
It demonstrates the effectiveness of the designed DCAF block.


\begin{figure}[t]
  \centering
  \includegraphics[width=0.45\linewidth]{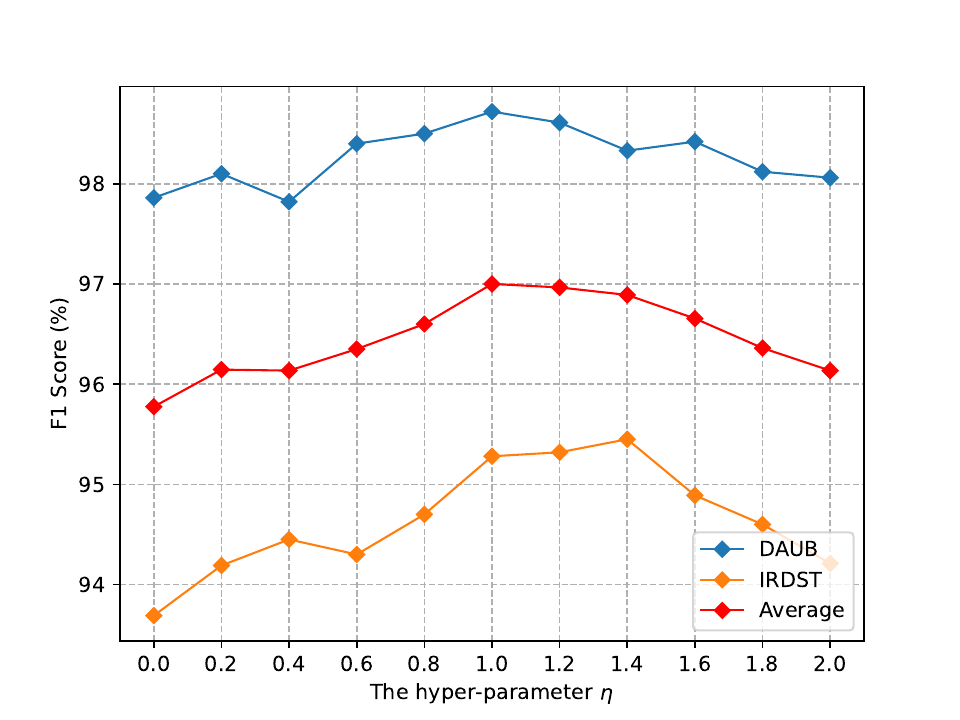}
  \caption{The impact of hyper-parameter $\eta$  in the loss function on  two datasets.}
  \label{eta}
\end{figure}

\textbf{Effectiveness of motion compensation loss $\boldsymbol{ \mathcal{L}_{MC} }$.}
Here, we show the necessity of motion compensation loss $\mathcal{L}_{MC}$. 
From the results in the 5th and 6th rows of Table \ref{Ablation}, it can be seen that although there is feature alignment, without the supervision of motion compensation loss, the performance is still sub-optimal.
 One possible reason is that deformable convolution is inherently difficult to train, and training instability often leads to offset overflow, deteriorating the final performance \cite{Chan2021UnderstandingDA}.
Therefore, it is necessary to use motion compensation loss for temporal alignment supervision.
Besides, we also investigate the impact of hyper-parameter $\eta$ in the loss function.
Fig. \ref{eta} shows
that the $F$1 score has a stable region that reaches a peak when $\eta=1.0$ on the DAUB dataset, and reaches a peak when $\eta=1.4$ on the IRDST dataset. 
In addition, we can also see that when $\eta=1.0$, the average $F$1 of the two datasets is optimal, so the hyper-parameter $\eta$ is  set to 1.0.

\textbf{Effectiveness of feature refinement module.}
We use the FR module to replace the simple multi-frame fusion operation (i.e., concatenation operation) to evaluate its effectiveness.
From the results in the 1st and 4th rows (or 3rd and 6th rows) in Table \ref{Ablation}, 
we can conclude that our feature refinement module contributes a lot to performance improvement.
For a deep investigation,  
we consider an additional ablation study about the branch configuration of the FR module. 
Intuitively, there are two variants of the FR module: 
FR without AFS (removing the Adaptive Fusion Structure (AFS) guided by the attention mechanism and using simple concatenation operation to fuse multiple features), 
and FR without AGDF (using 8 3 × 3 convolutional layers with 64 filters to replace the AGDF block).
Experimental results are presented in
Table \ref{Ablation_FR}. 
We can see that AFS and AGDF block are able to boost the performance of baseline, but all perform worse than the FR module.

\begin{table}[t]
  \centering
  \caption{Ablation Study on Feature Refinement Module.
  AFS: Adaptive Fusion Structure Guided by the
Attention Mechanism.}
  
    \begin{tabular}{cc|cc|cc}
    \hline
    
    \multirow{2}{*}{AFS} & \multirow{2}{*}{AGDF} & \multicolumn{2}{c|}{DAUB} & \multicolumn{2}{c}{IRDST} \\
\cline{3-6}
&       & mAP$_{50}$ (\%) & F1 (\%) & mAP$_{50}$ (\%) & F1 (\%) \\
    \hline
    
    -     & -     & 84.02  & 92.02  & 74.13  & 87.25  \\
    $\checkmark$     & -     & 86.81  & 93.51  & 76.88  & 88.63  \\
    -     & $\checkmark$     & 88.23  & 94.28  & 78.23  & 89.30  \\
    $\checkmark$     & $\checkmark$     & 91.92  & 96.56  & 80.88  & 90.45  \\
    \bottomrule
    \end{tabular}%
    
  \label{Ablation_FR}%
\end{table}%

\begin{figure}[t]
  \centering
  \includegraphics[width=0.6\linewidth]{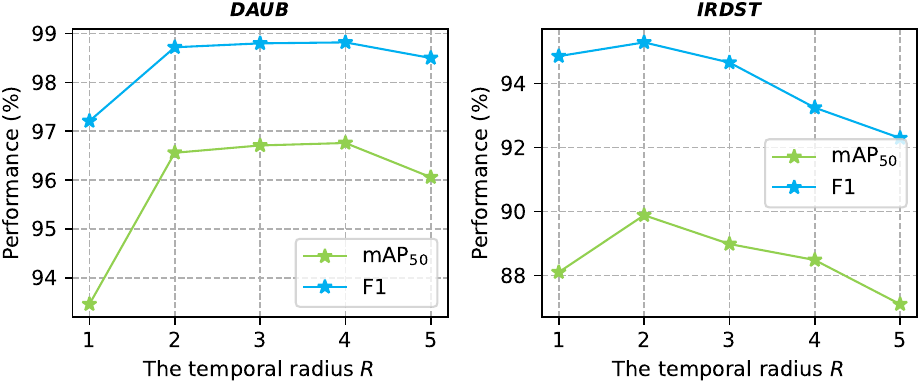}
  \caption{The impact of temporal radius $R$ on  two datasets.
  The performance is positively correlated with the temporal radius $R$, but reaches saturation
very quickly and even decreases.
  }
  \label{R}
\end{figure}

\subsubsection{Effectiveness of Temporal Radius \texorpdfstring{$R$}{}}
Although we follow SSTNet method \cite{chen2024sstnet} to use 5 frames  for model training, an ablation study is also conducted on the temporal radius $R$ to analyze its impact on performance. Here, due to limited memory,
we have only studied the case where the maximum value of
$R$ is 5.
As shown in Fig. \ref{R}, 
the performance of our DFAR method increases with the temporal radius $R$ in genera, but quickly reaches a plateau and even decreases when the number of frames is too large.
This phenomenon is more obvious in the IRDST dataset.
The reason is that when the temporal radius is too large, the greater motion makes it difficult for the model to be effectively aligned, and inaccurate alignment may even introduce new artifacts and degrade performance.
After comprehensively considering computation complexity and accuracy, we set $R=2$.

\section{Conclusion}
\label{Sec6}
We proposed a new end-to-end network for moving infrared dim-small target detection.
Specifically, a TDA module based on the designed DCAF block is proposed by explicitly aligning adjacent frames with the current frame at
the feature level to mine temporal information.
Furthermore, a feature
refinement module with the adaptive fusion structure and  AGDF blocks is designed to adaptively fuse and refine useful
temporal information from the aligned features.
In addition, we extend
the traditional loss function by introducing a new motion compensation loss to improve the temporal alignment.
Both qualitative and quantitative experimental results demonstrate the effectiveness of our proposed DFAR method, which can significantly improve detection performance compared with the state-of-the-art methods.
Ablation studies are also conducted to show  the effectiveness of different component assemblies in our method.




%




\bibliographystyle{IEEEtran}
\bibliography{References}

\newpage

\vfill

\end{document}